\documentclass[lettersize,journal]{IEEEtran}

\usepackage{amssymb}
\usepackage{amsmath,amsfonts}
\usepackage{bm}
\usepackage{algorithmic}
\usepackage{algorithm}
\usepackage{array}
\usepackage[caption=false,font=normalsize,labelfont=sf,textfont=sf]{subfig}
\usepackage{textcomp}
\usepackage{stfloats}
\usepackage{url}
\usepackage{verbatim}
\usepackage{graphicx}
\usepackage{cite}
\usepackage{multirow}
\usepackage{threeparttable}
\usepackage{booktabs}
\usepackage{bookmark}
\usepackage{graphicx}
\begin{document}

\title{SCA-PVNet: Self-and-Cross Attention Based Aggregation of Point Cloud and Multi-View for 3D Object Retrieval}

\author{Dongyun Lin, Yi Cheng, Aiyuan Guo, Shangbo Mao, Yiqun Li
\thanks{This research is supported by A*STAR under its RIE2020 INDUSTRY ALIGNMENT FUND - INDUSTRY
COLLABORATION PROJECTS (IAF-ICP) Grant No I2001E0073. (Corresponding author: Dongyun Lin)}

\thanks{All the authors are with Visual Intelligence Department, Institute for Infocomm Research, A*STAR, Singapore.}
}



\maketitle

\begin{abstract}
To address 3D object retrieval, substantial efforts have been made to generate highly discriminative descriptors of 3D objects represented by a single modality, e.g., voxels, point clouds or multi-view images. It is promising to leverage the complementary information from multi-modality representations of 3D objects to further improve retrieval performance. However, multi-modality 3D object retrieval is rarely developed and analyzed on large-scale datasets. In this paper, we propose self-and-cross attention based aggregation of point cloud and multi-view images (SCA-PVNet) for 3D object retrieval. With deep features extracted from point clouds and multi-view images, we design two types of feature aggregation modules, namely the In-Modality Aggregation Module (IMAM) and the Cross-Modality Aggregation Module (CMAM), for effective feature fusion. IMAM leverages a self-attention mechanism to aggregate multi-view features while CMAM exploits a cross-attention mechanism to interact point cloud features with multi-view features. The final descriptor of a 3D object for object retrieval can be obtained via concatenating the aggregated features from both modules. Extensive experiments and analysis are conducted on three datasets, ranging from small to large scale, to show the superiority of the proposed SCA-PVNet over the state-of-the-art methods.
\end{abstract}

\begin{IEEEkeywords}
Point Cloud, Multi-View, Cross-Attention Vision Transformer, Multi-Modal 3D Object Retrieval
\end{IEEEkeywords}

\section{Introduction}
\IEEEPARstart{R}{ecently}, a huge number of 3D data have been produced in many applications, ranging from autonomous driving to 3D printing. To manage the large-scale 3D data, an effective 3D object retrieval system is in demand, which is capable of generating a ranked list of objects according to the semantic similarity to a given query object.

In the community of computer vision and multimedia, 3D object retrieval has drawn significant research interest by exploiting different modalities of 3D objects, such as voxel, point cloud and multi-view images. The existing methods for 3D object retrieval are mainly deep learning models designed based on a single modality. For example, VoxNet~\cite{maturana2015voxnet}, DGCNN~\cite{yue2019dynamic} and MVCNN~\cite{su2015multi} are typical models based on voxel, point cloud and multi-view representations of 3D objects, respectively. Within the category of unimodal methods, voxel-based approaches acquire descriptors of 3D objects by employing dense and regular grids formed by voxels. Although voxel-based methods have the advantage of leveraging the entirety of object information, they often encounter challenges related to high computational complexity. On the other hand, point-cloud-based methods obtain descriptors of 3D objects by utilizing a set of sampled 3D points extracted from the objects. Point-cloud-based methods can also leverage almost the complete information of the objects if the sampled points are dense and they can generally reduce the computation complexity comparing with voxel-based-methods. Multi-view-based methods learn the descriptors of 3D objects via two steps: (i) rendering several view images by projecting the 3D object from multiple viewpoints; (ii) aggregating deep visual features extracted from every view image into a unified descriptor. Along with the advancement of well-engineered 2D deep CNN networks for discriminative feature extraction, multi-view-based methods have achieved the state-of-the-art retrieval performance.

\begin{figure}[!t]
  \centering
  \vspace{8mm}
  \includegraphics[scale=0.35]{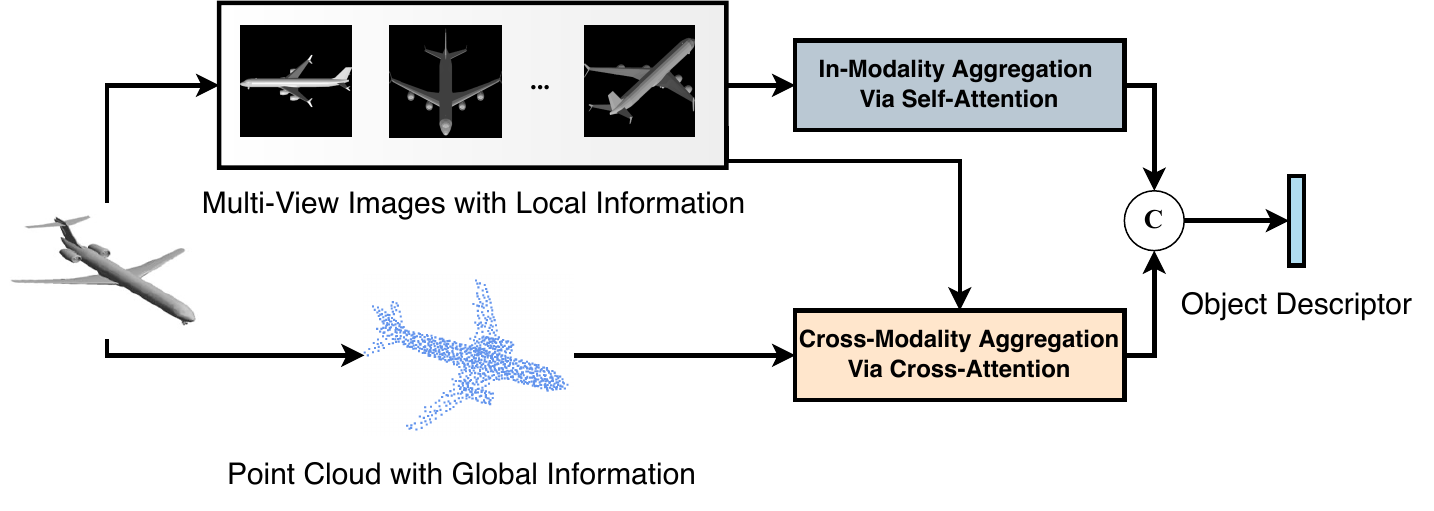}
  \caption{The illustration of our motivation of jointly leveraging the local information and global information of the 3D object from its multi-view and point cloud representations via the proposed modules.}
  \label{fig:Motivation}
\end{figure}

Based on the aforementioned analysis, the multi-view-based methods (e.g., MVCNN) only exploit local and partial information of 3D objects via a limited number of projected view images while point-cloud-based methods (e.g., DGCNN) can efficiently learn global representations of 3D objects via exploiting the complete information of the objects. Hence, it is promising to leverage the complementary information of multi-modality representations of 3D objects to generate more discriminative descriptors and thereby improve retrieval performance. There are several methods proposed for multi-modality 3D object retrieval, such as PVNet~\cite{you2018pvnet}, PVRNet~\cite{you2019pvrnet} and MMFN~\cite{nie2020mmfn}. These methods focus on how to effectively fuse point cloud and multi-view features. Also, most existing works are  evaluated only using small-scale datasets like ModelNet40 but rarely analyzed on large-scale datasets.

Motivated by these findings, in this paper, we propose self-and-cross attention based aggregation of point cloud and multi-view images (SCA-PVNet) for 3D object retrieval. It consists of two branches for point cloud modality and multi-view modality inputs, respectively. Within the multi-view modality branch, we further exploit two types of representations of multi-view images in two hierarchical levels, namely object-level and view-level features, respectively. Object-level features are generated by fusing the intermediate CNN features for all the view images while the view-level features are generated by average-pooling each of the intermediate CNN features for each view image. For both object-level and view-level features, to achieve single-modality and cross-modality feature aggregation, we propose two types of feature aggregation modules, namely In-Modality Aggregation Module (IMAM) and Cross-Modality Aggregation Module (CMAM). IMAM leverages self-attention mechanism to aggregate the multi-view features. CMAM exploits cross-attention mechanism to interact the point cloud feature with the aggregation process of multi-view features. Finally, we concatenate the object-level and view-level aggregated features output by IMAM and CMAM and feed the concatenated feature through a fully-connected layer to obtain the final descriptors for object retrieval. Our motivation of jointly leveraging the local information and global information from multi-view and point cloud modality is illustrated in Fig.~\ref{fig:Motivation}.


The major contributions of this paper are summarized as below:

\begin{itemize}
   \item We propose self-and-cross attention based aggregation of point cloud and multi-view images (SCA-PVNet) for 3D object retrieval. This network enhances the aggregation process of multi-view features under two hierarchical levels, namely object-level and view-level features.

   \item We propose the In-Modality Aggregation Module (IMAM) and the Cross-Modality Aggregation Module (CMAM) to facilitate in-modality and cross-modality feature aggregation, respectively.

   \item We evaluate and analyze the proposed SCA-PVNet on three datasets (ModelNet40, ShapeNetCore55 and MCB-A), ranging from small to large scale, to show the superiority of the proposed method over the state-of-art methods.

\end{itemize}

\begin{figure*}[!t]
  \centering
  \includegraphics[scale=0.35]{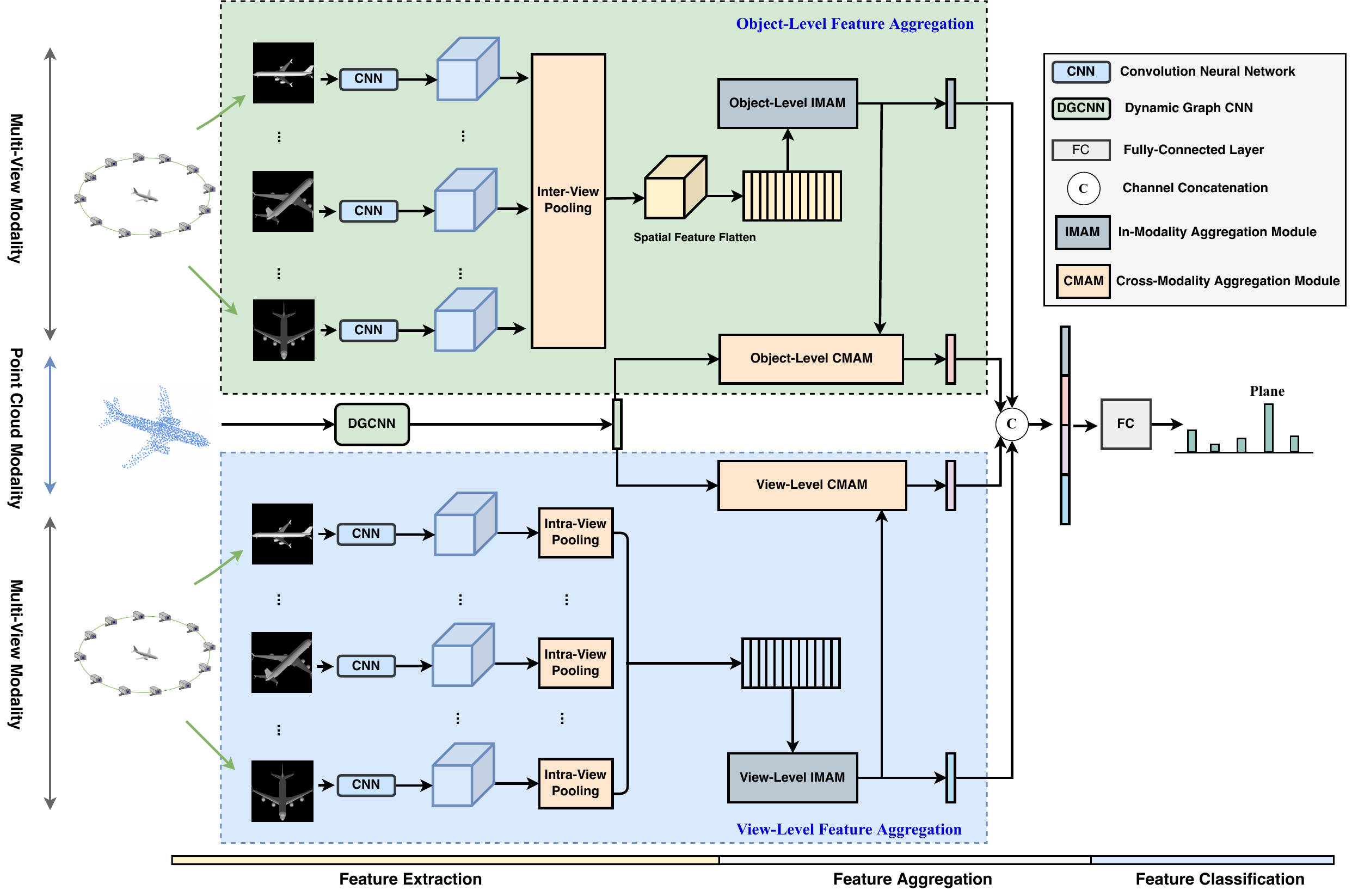}
  \caption{The overall architecture of the proposed SCA-PVNet.}
  \label{OverallArchitecture}
\end{figure*}

\section{Related Work}
In this section, we provide a concise review of methods for 3D object retrieval, specifically focusing on uni-modal-based methods and multi-modal-based methods.

\subsection{Uni-Modal-Based Methods}
Uni-model-based methods can be divided into three categories: voxel-based-methods, point-cloud-based methods and multi-view-based methods. Voxel-based-methods learn the descriptors of 3D objects represented by voxelized grids using 3D deep learning models~\cite{wu20153d,maturana2015voxnet,brock2016generative}. For example, in 3D ShapeNet~\cite{wu20153d}, a deep belief network was employed to learn feature representations of 3D objects. Another example is VoxNet~\cite{maturana2015voxnet}, a 3D convolutional neural network designed to learn feature representations of voxelized grids. While voxel-based methods have the advantage of utilizing the complete information of 3D objects, they often encounter challenges related to significant computational complexity. Point-cloud-based methods learn the representations of objects from the points uniformly sampled from the object surfaces~\cite{qi2017pointnet,QiPointNet++,yue2019dynamic,cheraghian20193dcapsule}. The pioneering work PointNet~\cite{qi2017pointnet} was designed to build up a 3D deep network to classify 3D objects from their point cloud representations. PointNet++~\cite{QiPointNet++} was designed to exploit hierarchical multi-scale features of PointNet. 3DCapsule~\cite{cheraghian20193dcapsule} introduced a Capsule network into PointNet architecture for point cloud classification. In general, the computational complexity of point-cloud-based methods is lower compared to that of voxel-based methods. Multi-view methods render several view images via projecting a 3D object from multiple viewpoints and learn the descriptors by aggregating the visual features extracted from these view images~\cite{su2015multi,he2020improved,zhou2019multi,nie2021dan,lin2022multi,xu2019enhancing,han20193d2seqviews,he2018triplet,li2019angular,ma2018learning,liu2021hierarchical,han2018seqviews2seqlabels,kanezaki2019rotationnet,wei2020view,he2019view,su2018deeper,nie20203d,jiang2019mlvcnn}. The main stream of research works represented by MVCNN~\cite{su2015multi} and its variants~\cite{he2020improved,lin2022multi,wei2020view,he2018triplet,li2019angular,zhou2019multi,nie2021dan,xu2019enhancing,liu2021hierarchical,kanezaki2019rotationnet,wei2020view,he2019view,su2018deeper,jiang2019mlvcnn} applied well-engineered deep CNNs to extract the features from the view images and fuse these features into the descriptors of 3D objects. Another group of works applied RNNs to aggregate the deep features extracted from each view images based on the assumption that the view images as a sequence of temporally correlated views~\cite{han20193d2seqviews,ma2018learning,han2018seqviews2seqlabels}. Thanks to the advancement of CNN/RNN networks, multi-view-based methods have achieved the state-of-the-art performance for 3D object retrieval. Even though uni-modal-based methods have exhibited such good performance, it is promising to explore and leverage the complementary information from various data modalities to further improve the retrieval performance. In this work, we tackle the task of multi-modal 3D object retrieval.

\subsection{Multi-Modal-Based Methods}
Recently, there are several methods proposed for multi-modality 3D object retrieval. Since learning from voxelized dense grids is computationally expensive, most existing works focus on leveraging point cloud and multi-view representations of 3D objects to learn discriminative descriptors for retrieval. PVNet~\cite{you2018pvnet} was a trailblazing initiative that combined point cloud and multi-view data for the purpose of 3D object recognition. The model made use of high-level features from multi-view data to analyze and discern the correlation and distinctiveness of structural features derived from point cloud data. PVRNet~\cite{you2019pvrnet} was proposed to effectively fuse point cloud and multi-view features with a proposed relation score module. MMFN~\cite{nie2020mmfn} leveraged the correlations among point cloud, multi-view and panorama-view modality by introducing a correlation loss and an instance loss. In~\cite{peng2020attention}, a novel attention-guided fusion network of point cloud and multi-views was proposed for 3D object recognition. Compared with these existing works focusing on how to effectively fuse multi-modality features, our SCA-PVNet tackles how to correlate and interact cross-modality features during feature aggregation process. Also, the existing methods were rarely evaluated on large-scale datasets, in this work, we evaluate the proposed SCA-PVNet on three datasets which range from small to large scale to show the superiority of the proposed method.

\section{Proposed Method}

In this section, we introduce the proposed network for multi-modality 3D object retrieval. The overall architecture of the
proposed network is shown in Fig.~\ref{OverallArchitecture}. In general, it consists of three stages: (i) Feature Extraction; (ii) Feature Aggregation and (iii) Feature Classification. The network takes two types of modalities as input: point clouds and
multi-view images, respectively. Given a 3D object, it is firstly represented as a point cloud and a set of view images captured from multiple viewpoints. At the feature extraction stage, the point cloud is fed through DGCNN~\cite{yue2019dynamic} to obtain the point cloud feature while the view images are fed into a CNN network to generate the deep feature representations. For view feature extraction, we exploit two hierarchical levels of representations, i.e., object-level and view-level features. At the feature aggregation stage, firstly,
the object-level features and the view-level features are aggregated via a self-attention process to generate an object-level global feature and a view-level global feature, respectively. In parallel, we propose to leverage cross-attention mechanism to correlate
the point cloud feature with the object-level and view-level features to obtain the object-level and view-level cross-modality global features, respectively. Finally, at the feature classification stage, we concatenate all the global features and exploit a fully-connected layer to map the concatenated feature to the final descriptor. The details of each stage are described in the following subsections.


\subsection{Feature Extraction}
In this work, we exploit two modalities to represent a 3D object, i.e., point clouds and multi-view images as shown in Fig.~\ref{OverallArchitecture}. For point cloud modality, the object $\mathcal{O}_i$ is represented by $n$ sampled points via farthest point sampling (FPS) algorithm, denoted by $P_i=\{\bm{p}_1, \bm{p}_2, \ldots, \bm{p}_n\}$, where $\bm{p}_j\in\mathbb{R}^3$ contains 3D coordinates of the $j^{th}$ point. DGCNN is exploited to extract the vectorized representation for the point cloud modality. In accordance with the model outlined in \cite{yue2019dynamic}, the DGCNN is built using a 3D spatial transform network, several EdgeConv layers, and a max pooling operation. ${\bm{P}}_i$ is fed through DGCNN to generate the point cloud modality feature:
\begin{equation}
  {\bm{f}}_{{\rm{point}},i} = {\rm{DGCNN}}({\bm{P}}_i; \theta_{\rm{point}}),
  \label{eq:DGCNN_feature}
\end{equation}
where ${\bm{f}}_{{\rm{point}},i}$ denotes the point cloud modality feature for $\mathcal{O}_i$ and $\theta_{\rm{point}}$ denotes the learnable parameters for the DGCNN.

For multi-view representation of object $\mathcal{O}_i$, multiple view images $\{{\bm{I}}^1_i, {\bm{I}}^2_i, \ldots, {\bm{I}}^M_i\}$ are rendered by setting up $M$ virtual cameras around the object with the uniform interval angle of 30 degrees along the Z axis. The multi-view modality features are generated by feeding these view images through a CNN without the pooling operation and the classification head:

\begin{equation}
  {\bm{F}}^{j}_{{\rm{view}},i} = {\rm{CNN}}({\bm{I}}^j_i;\theta_{\rm{view}}),\  j=1,2,\ldots,M,
\end{equation}
where ${\bm{F}}^{j}_{{\rm{view}},i}\in \mathbb{R}^{C \times H \times W}$ denotes the feature map for the $j_{th}$ view of $\mathcal{O}_i$ and $\theta_{\rm{view}}$ denotes parameters of the CNN. With these intermediate CNN feature maps corresponding to the view images, we introduce two hierarchical levels of representations for multi-view modality, i.e., object-level and view-level representations. To generate object-level representation $\bm{F}_{\rm{object},i}$, the intermediate feature maps $\{{\bm{F}}^{1}_{{\rm{view}},i}, {\bm{F}}^{2}_{{\rm{view}},i}, \ldots, {\bm{F}}^{M}_{{\rm{view}},i}\}$ are fed through an Inter-View Pooling module where the feature maps are fused via average pooling across the views:
\begin{equation}
  {\bm{F}}_{{\rm{object}},i} = {\rm{AveragePool}}({\bm{F}}^{1}_{{\rm{view}},i}, {\bm{F}}^{2}_{{\rm{view}},i}, \ldots, {\bm{F}}^{M}_{{\rm{view}},i}).
\end{equation}
${\bm{F}}_{{\rm{object}},i}\in\mathbb{R}^{C \times H \times W}$ summarizes the multi-view images and it consists of $HW$ $C-$dimensional vectorized features $\{{\bm{f}}^{l}_{{\rm{object}},i}\}(l=1,2,\ldots, HW)$. ${\bm{f}}^{l}_{{\rm{object}},i}$ contains the global object-level semantics of $\mathcal{O}_i$ at location $l$. In addition to object-level view representation, we also introduce view-level representation by feeding each intermediate feature map through an Intra-View Pooling module via average pooling operation:
\begin{equation}
  {\bm{f}}^{j}_{{\rm{view}},i} = {\rm{AveragePool}}({\bm{F}}^{j}_{{\rm{view}}, i})\ j=1,2,\ldots,M.
\end{equation}
Here, ${\bm{f}}^j_{{\rm{view}},i}$ contains the local view-level semantics of $\mathcal{O}_i$ reflected in view image ${\bm{I}}^j$.

\subsection{Feature Aggregation}
After feature extraction stage, given $\mathcal{O}_i$, for the point cloud modality, the global feature ${\bm{f}}_{{\rm{point}},i}$ is generated. For the multi-view modality, we obtain a set of object-level global features $\{{\bm{f}}^{l}_{{\rm{object}},i}\}(l=1,2,\ldots, HW)$ and a set of view-level local features $\{{\bm{f}}^{j}_{{\rm{view}},i}\}(j=1,2,\ldots, M)$, respectively.

At feature aggregation stage, we design two types of feature aggregation modules: In-Modality Aggregation Module ($\bf{IMAM}$) and Cross-Modality Aggregation Module ($\bf{CMAM}$). On one hand, IMAM leverages the multi-head \textbf{self-attention} mechanism inspired by Vision Transformer (ViT)~\cite{dosovitskiy2020image} to aggregate object-level global features and view-level local features, respectively. On the other hand, CMAM exploits \textbf{cross-attention} mechanism inspired by Cross-Attention Vision Transformer (CrossVit)~\cite{chen2021crossvit} to interact the point cloud global feature with the object-level and view-level features in the aggregation process and thereby to generate cross-modality global representations of the object.

\begin{figure}[!t]
  \centering
  \includegraphics[scale=0.4]{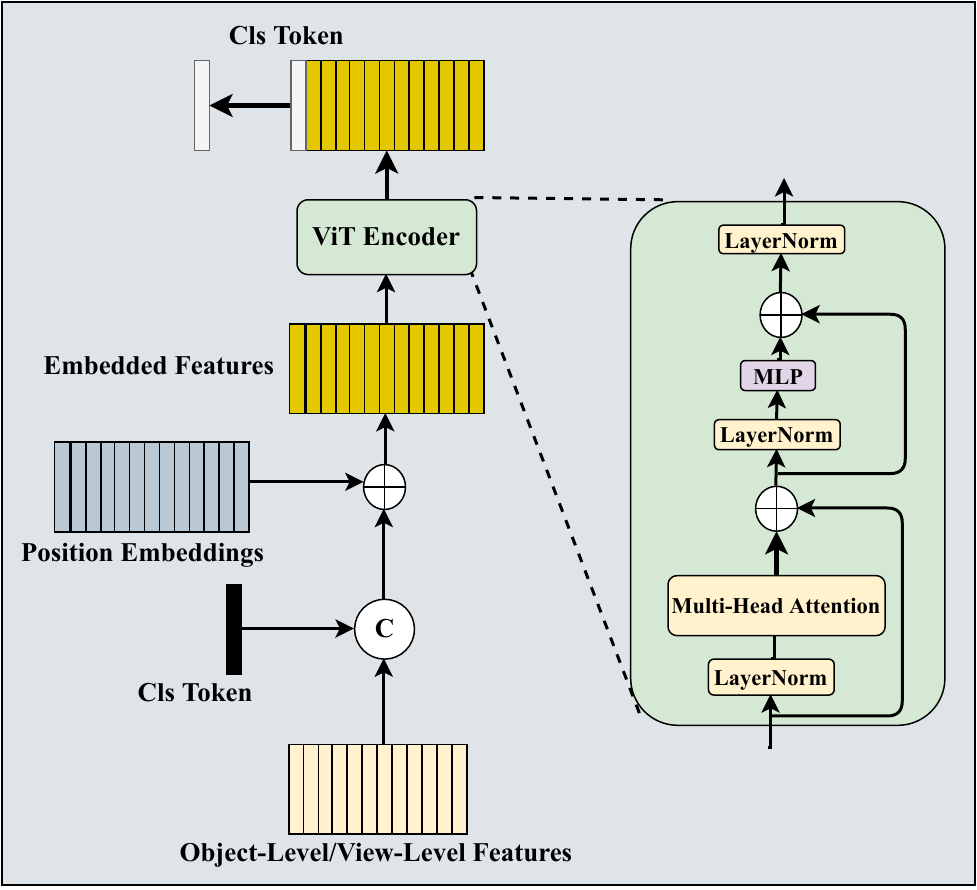}
  \caption{The architecture of IMAM.}
  \label{fig:IMAM}
\end{figure}

\subsubsection{\textbf{In-Modality Aggregation Module}}\label{sec:IMAM}
As shown in Fig.~\ref{OverallArchitecture}, two IMAMs are proposed corresponding to the object-level global features and the view-level local features, respectively, i.e., object-level IMAM and view-level IMAM. For simplicity, we denote the two levels of input features for $\mathcal{O}_i$ as $\bm{F}_{{\rm{T}},i}$, where $\rm{T} \in \{object, view\}$. Given $\bm{F}_{{\rm{T}},i}$ as the input, a class token is prepended to $\bm{F}_{{\rm{T}},i}$ which is learnable, serving as the summarizing feature representation for the input features. Then, a set of learnable position embeddings is added to the input features (together with the class token feature) to generate the position relevant feature embeddings $\bm{Z}_{{\rm{T}},i}$. To aggregate the features, as illustrated in Fig.~\ref{fig:IMAM}, $\bm{Z}_{{\rm{T}},i}$ is fed through a Vision Transformer (ViT) Encoder:
\begin{align}
  \bm{Z}^{'}_{{\rm{T}},i} &= {\rm{MSA}}({\rm{LN}}(\bm{Z}_{{\rm{T}},i})) + \bm{Z}_{{\rm{T}},i}, \\
  \bm{Z}^{*}_{{\rm{T}},i} &= {\rm{LN}}({\rm{MLP}}({\rm{LN}}(\bm{Z}^{'}_{{\rm{T}},i})) + \bm{Z}^{'}_{{\rm{T}},i}),
\end{align}
where $\rm{LN}$ denotes LayerNorm operation, $\rm{MSA}$ denotes MultiHead Self-Attention (MSA) and $\rm{MLP}$ refers to MultiLayer Perceptron (MLP) blocks. After feature aggregation process, the object-level and view-level self-attended global features denoted by $\bm{f}^S_{{\rm{object}},i}$ and $\bm{f}^S_{{\rm{view}},i}$ are the features associated with the class token:
\begin{equation}
 \bm{f}^S_{{\rm{T}},i} = \bm{Z}^{*}_{{\rm{T}},i}[\rm{class\ token}]\quad {\rm{T}} \in \{\rm{object}, \rm{view}\}.
\end{equation}

\begin{figure}[htbp]
  \centering
  \includegraphics[scale=0.4]{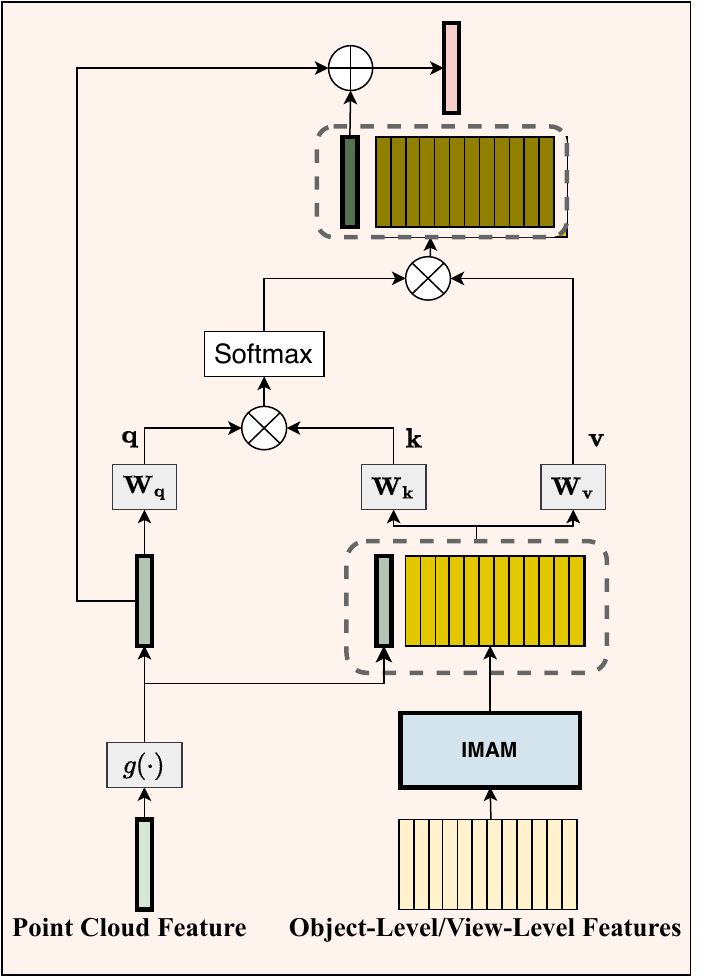}
  \caption{The architecture of CMAM.}
  \label{fig:CMAM}
\end{figure}

\subsubsection{\textbf{Cross-Modality Aggregation Module}}
Similar to IMAM, as shown in Fig.~\ref{fig:CMAM}, two CMAMs are proposed to implement cross-modality aggregation for object-level global features and view-level local features, respectively, i.e., object-level IMAM and view-level IMAM. The key motivation of CMAM is correlating the point cloud modality global feature to the aggregation process of multi-view modality features to enhance the feature discriminative capability. Towards this motivation, we leverage the cross-attention mechanism introduced in Cross-Attention ViT to design the CMAMs. Specifically, given $\bm{F}_{{\rm{T}},i}$ for $\mathcal{O}_i$ where $\rm{T} \in \{object, view\}$ as the input, the input features are first fed through an IMAM module to generate the aggregated features as introduced in Section~\ref{sec:IMAM}:
\begin{equation}
  \bm{Z}^S_{{\rm{T}},i} = {\rm{IMAM}}(\bm{F}_{{\rm{T}},i}).
\end{equation}
Subsequently, the global feature for point cloud modality $\bm{f}_{\rm{point}}$ generated from Eq. (\ref{eq:DGCNN_feature}) is concatenated with the aggregated multi-view features $\bm{Z}^S_{\rm{T}}$ to construct the hybrid aggregated features $\bm{Z}^H_{\rm{T}}$:
\begin{equation}
  \bm{Z}^H_{{\rm{T}},i} = g(\bm{f}_{{\rm{point}},i}) \oplus \bm{Z}^S_{{\rm{T}},i},
\end{equation}
where function $g(\cdot)$ performs feature dimension alignment and $\oplus$ denotes feature concatenation. Finally, to obtain the cross-modality global feature $\bm{f}^C_{{\rm{T}},i}$, the point cloud feature $g(\bm{f}_{{\rm{point}},i})$ serves as the query vector to interact with the hybrid aggregated features $\bm{Z}^H_{{\rm{T}},i}$ via a cross-attention mechanism:
\begin{align}
{\bm{q}} &= g({\bm{f}}_{{\rm{point}},i}){\bm{W}}_q, \ {\bm{k}} = {\bm{Z}}^H_{{\rm{T}},i}{\bm{W}}_k,\  {\bm{v}} = {\bm{Z}}^H_{{\rm{T}},i}{\bm{W}}_v, \\
{\bm{A}} &= {\rm{softmax}}({\bm{q}}{\bm{k}}^T / \sqrt{D/h}), \\
{\bm{Z}}^C_{{\rm{T}},i} &= {\bm{A}}{\bm{v}},
\end{align}
where ${\bm{W}}_q$, ${\bm{W}}_k$ and ${\bm{W}}_v$ refer to the learnable transformation matrices to generate query, key and value features for the attention computation. $D$ denotes the feature dimension and $h$ denotes the number of heads for MultiHead Self-Attention module. As shown in Fig.~\ref{fig:CMAM}, the object-level and view-level cross-modality global features ${\bm{f}}^C_{{\rm{object}},i}$ and ${\bm{f}}^C_{{\rm{view}},i}$ are calculated by adding $g({\bm{f}}_{{\rm{point}},i})$ with the class token feature in ${\bm{Z}}^C_{{\rm{T}},i}$:
\begin{equation}
{\bm{f}}^C_{{\rm{T}},i} = g({\bm{f}}_{{\rm{point}},i}) + {\bm{Z}}^C_{{\rm{T}},i}[{\rm{class}}\ {\rm{token}}]\quad {\rm{T}} \in \{\rm{object}, \rm{view}\}.
\end{equation}

\subsection{Feature Classification}
After feature aggregation stage, for $\mathcal{O}_i$, four aggregated features are obtained, namely object-level and view-level self-attended global features $\bm{f}^S_{{\rm{object}},i}$ and $\bm{f}^S_{{\rm{view}},i}$, together with object-level and view-level cross-modality global features ${\bm{f}}^C_{{\rm{object}},i}$ and ${\bm{f}}^C_{{\rm{view}},i}$. The final descriptor ${\bm{f}}_{{\rm{global}},i}$ is obtained by feeding forward the concatenation of all the four aggregated features through a MLP layer:
\begin{equation}
  {\bm{f}}_{{\rm{global}},i} = {\rm{MLP}}(\bm{f}^S_{{\rm{object}},i} \oplus \bm{f}^S_{{\rm{view}},i} \oplus {\bm{f}}^C_{{\rm{object}},i} \oplus {\bm{f}}^C_{{\rm{view}},i}).
\end{equation}
To generate a highly discriminative final descriptor for object retrieval, the Additive Angular Margin loss (ArcFace)~\cite{deng2019arcface} is exploited to train the network. Given a mini-batch of N objects belonging to K classes, we first generate the final descriptors ${\bm{f}}_{{\rm{global}},i}$ ($i=1,2,\ldots, N$). These final descriptors are normalized using their L2 norm, i.e., ${\bm{\hat{f}}}_{{\rm{global}},i} = \frac{{\bm{f}}_{{\rm{global}},i}}{||{\bm{f}}_{{\rm{global}},i}||}$. Then, we introduce a weight matrix $\bm{W} \in \mathbb{R}^{D \times K}$ where the $k^{th}$ column of $\bm{W}$ denoted by ${\bm{W}}_k$ is related to the $k^{th}$ category and is also L2 normalized. To enhance the discriminative capability of features, the ArcFace loss incorporates an additional angular margin to the angle beween the feature and the weight associated with the ground truth class:
\begin{equation}
{\mathcal{L}}=-\frac{1}{N}\sum_{i=1}^{N}\log\frac{e^{s\cos(\theta_{y_i} + m)}}{e^{s\cos(\theta_{y_i} + m)}+\sum_{k=1,k\neq  y_i}^{K}e^{s\cos\theta_{k}}},
\end{equation}
where $y_i$ denotes the ground truth class of the $i^{th}$ training sample. $\theta_{k}\triangleq\arccos({\bm{W}}^{{\rm{T}}}_k{\bm{\hat{f}}}_{{\rm{global}},i})$ denotes the angle between ${\bm{W}}_k$ and ${\bm{\hat{f}}}_{{\rm{global}},i}$. $m$ represents the value of the additive angular margin, while $s$ represents the scaling factor. The overall network is trained by minimizing the ArcFace loss $\mathcal{L}$. After training, the final descriptor of every object is adopted for 3D object retrieval.

\section{Experiments}
In this section, the experiments and discussions are conducted on three benchmarking datasets, ranging from small to large scale, to show the effectiveness of the proposed SCA-PVNet.
\subsection{Datasets and Evaluation Metrics}
We adopt three benchmarking datasets to evaluate the retrieval performance, i.e., ModelNet40, ShapeNetCore55 and Mechanical Components Benchmark (MCB). The ModelNet40 dataset comprises 12,311 3D CAD models derived from 40 distinct categories, partitioned into a training set with 9,843 objects and a testing set with 2,468 objects. ShapeNetCore55 includes 51,162 3D objects categorized into 55 primary classes and 203 subordinate classes. Following the official setting in SHREC2017 competition, the training/validation/testing splitting is 70\%/10\%/20\%. We leverage the aligned version of the ShapeNetCore55 dataset, which consists of organized 3D objects. The MCB dataset consists of 58,696 objects across 68 categories. For our experiments, we utilize the MCB dataset's version A (MCB-A) wherein the objects are uniformly aligned, adhering to the training and testing split configuration as established in the original study.~\cite{kim2020large}.

We generated the point cloud representation of each 3D object by sampling 1024 points using the farthest point sampling (FPS) method. Additionally, we rendered 12 view images of the object using Blender software and Phong shading. This was accomplished by positioning 12 virtual cameras around the object, each at a 30-degree angle along the Z-axis.

For the evaluating metrics on ModelNet40, we utilize Mean Average Precision (MAP). This involves calculating the average precision for each 3D object in the test set, using it as a query sample. A ranked list of all the test objects is then generated based on pair-wise cosine distances, and the mean of these average precisions is calculated. For ShapeNetCore55, we use the SHREC2017 competition released evaluator to compute the retrieval performance in terms of F1@N, mAP and Normalized Discounted Cumulative Gain (NDCG) under ``micro'' and ``macro'' settings, respectively. For MCB dataset, we report F1@N, mAP and NDCG@N under ``micro'', ``macro'' and ``micro + macro'' settings, respectively.



\subsection{Implementation and Training Details}
At feature extraction stage, DGCNN is adopted for point cloud modality while MVCNN is for multi-view modality. We exploit AlexNet as the backbone network of MVCNN for ModelNet40 and ResNet-34 for ShapeNetCore55 and MCB-A. Following ~\cite{you2019pvrnet}, both DGCNN and MVCNN are pretrained on the training set using cross-entropy loss. During the feature aggregation phase, for the IMAM model, we set the multi-head self-attention count to 4. Additionally, the dimension of the hidden feature is set to 128, while the dimension of the aggregated features is set to 512. For CMAM, the number of attention head is set as 4, the dimension of aggregated feature is set as 512. At feature classification stage, the fully connected layer maps the concatenated aggregated feature ${\bm{f}}_{{\rm{global}},i}$ into a vector whose dimension is the same as the number of object categories. In the pretraining phase, the batch size, learning rate, weight decay, and number of epochs are respectively set at 400, 0.01, 0.001, and 50. For fine-tuning, we follow the settings in PVRNet~\cite{you2019pvrnet}, during the initial 10 epochs, the feature extraction models remain unchanged while the remaining components, namely IMAM, CMAM, and the FC layers are fine-tuned. The learning rate for the first 10 epoch is set as 0.01 while 0.001 after 10 epochs. The batch size is set as 28 per GPU, the weight decay is set as $1e^{-5}$ and the number of epoch is set as 50. For ArcFace loss, the angular margin is set as 0.5 and the scale is set as 64.

\subsection{Comparison with State-of-the-art Methods}

\subsubsection{\textbf{Performance Comparison on ModelNet40 Dataset}}
Table~\ref{tab:comparsions_on_modelnet40} shows the retrieval performance on ModelNet40 in mAP. The table indicates that our SCA-PVNet outperforms all other methods compared in the experiment. In particular, our method outperforms all the methods (PVNet, PVRNet, MMFN and Attention-Guided FUsion Network) which use point cloud and multi-view modalities for 3D object retrieval. Compared with PVRNet which uses the same feature extraction networks (DGCNN and MVCNN with AlexNet) as our method, our method produces 2.0\% performance gain in mAP, reflecting the effectiveness of the proposed IMAM and CMAM in feature aggregation. It is also worth noting that compared with MMFN which use three data modalities including point cloud, multi-view and panorama views, our method can still achieves better performance (with 2.2\% performance gain) even though we only exploit two data modalities. In addition, our method outperforms those methods based on single modality (either point cloud or multi-view). Particularly, compared with vanilla MVCNN~\cite{su2015multi}, we produce 12.3\% performance gain. Compared with the advanced MVCNN variants, such as MVCNN with cube loss~\cite{wang2019learning}, we still produce 3.6\% performance gain. These observations show the superiority of effectively incorporating point cloud representation and interacting it with multi-view representations. Finally, it is evident that our model, trained using ArcFace loss, outperforms models trained with other metric learning losses like TCL~\cite{he2018triplet,he2020improved}, ATCL~\cite{li2019angular}, and cube loss~\cite{wang2019learning}.

\begin{table}[htbp]
\caption{Performance comparison on ModelNet40 (in \%). The best results are in bold.}
\centering
{\scalebox{0.8}{
\begin{tabular}{|l*{1}{p{0.94cm}<{\centering}}|}
\hline
\multirow{1}{*}{Methods} & \multicolumn{1}{c|}{mAP} \\
\hline
SPH~\cite{kazhdan2003rotation} &33.3\\
LFD~\cite{chen2003visual} &40.9  \\
3DShapeNet~\cite{wu20153d} &49.2 \\
DLAN~\cite{furuya2016deep} &85.0 \\
DeepPano~\cite{shi2015deeppano} &76.8 \\
GIFT~\cite{bai2016gift} &81.9 \\
MVCNN~\cite{su2015multi} &80.2 \\
RED~\cite{bai2017ensemble} & 86.3 \\
GVCNN~\cite{feng2018gvcnn} &85.7 \\
TCL~\cite{he2018triplet} &88.0 \\
ATCL~\cite{li2019angular} &86.1 \\
SeqViews2SeqLabels~\cite{han2018seqviews2seqlabels} & 89.1 \\
VDN~\cite{leng2018learning} &86.6 \\
Batch-wise~\cite{xu2019learning} &83.8 \\
MVCNN (Cube loss)~\cite{wang2019learning} & 88.9 \\
VNN~\cite{he2019view} & 88.9 \\
NCENet~\cite{xu2019enhancing} &87.1 \\
3DViewGraph~\cite{han20193dviewgraph} & 90.5 \\
GSL (ATCL)~\cite{he2020improved} & 90.1 \\
DAN~\cite{nie2021dan} & 90.4 \\
DAGL+Attention-NetVlad~\cite{shi20223d}& 91.1 \\
PVNet~\cite{you2018pvnet} & 89.5 \\
PVRNet~\cite{you2019pvrnet} & 90.5 \\
MMFN~\cite{nie2020mmfn} & 90.3 \\
Attention-Guided Fusion Network~\cite{peng2020attention} & 91.4 \\
\hline
Our SCA-PVNet &\textbf{92.5} \\ \hline
\end{tabular}
}}
\label{tab:comparsions_on_modelnet40}
\end{table}

\subsubsection{\textbf{Performance Comparison on ShapeNetCore55 Dataset}}
Table~\ref{tab:comparison_on_shapenet} shows the retrieval performance on ShapeNetCore55 dataset. Based on the table, it can be noted that our SCA-PVNet demonstrates superior performance in 5 out of 6 performance metrics, excluding F1@N. While RotationNet emerged as the winning solution in the SHREC2017 competition for 3D object retrieval, our method outperforms it in terms of mAP and NDCG in both micro and macro settings. For F1@N, our method achieves the same performance as RotationNet under micro setting and only 0.8\% lower under macro setting. Compared with MVCNN, our method achieves significant performance gain in all 6 metrics which shows the effectiveness of incorporating additional point cloud modality.

\begin{table}[htbp]
  \centering
  \caption{Performance comparison on ShapeNetCore55 (in \%)}
  {\scalebox{0.8}{
    \begin{tabular}{|c|ccc|ccc|}
    \toprule
    \multirow{2}[2]{*}{Methods} & \multicolumn{3}{c|}{micro} & \multicolumn{3}{c|}{macro} \\
          & F1@N  & mAP   & NDCG  & F1@N  & mAP   & NDCG \\
    \midrule
    RotationNet~\cite{kanezaki2019rotationnet} & \textbf{79.8} & 77.2  & 86.5  & \textbf{59.0} & 58.3  & 65.6 \\
    Improved\_GIFT~\cite{bai2017gift} & 76.7  & 72.2  & 82.7  & 58.1  & 57.5  & 65.7 \\
    ReVGG~\cite{bai2015beyond} & 77.2  & 74.9  & 82.8  & 51.9  & 49.6  & 55.9 \\
    DLAN~\cite{furuya2016deep}  & 71.2  & 66.3  & 76.2  & 50.5  & 47.7  & 56.3 \\
    MVFusionNet~\cite{daras2009compact} & 69.2  & 62.2  & 73.2  & 48.4  & 41.8  & 50.2 \\
    CM-VGG5-6DB~\cite{lian2013cm} & 47.9  & 54.0  & 65.4  & 16.6  & 33.9  & 40.4 \\
    ZFDR~\cite{li20133d} & 28.2  & 19.9  & 33.0  & 19.7  & 25.5  & 37.7 \\
    DeepVoxNet~\cite{maturana2015voxnet}  & 25.3  & 19.2  & 27.7  & 25.8  & 23.2  & 33.7 \\
    GIFT~\cite{bai2017gift}  & 68.9  & 64.0  & 76.5  & 45.4  & 44.7  & 54.8 \\
    MVCNN~\cite{su2015multi} & 76.4  & 73.5  & 81.5  & 57.5  & 56.6  & 64.0 \\
    \midrule
    Our SCA-PVNet  & \textbf{79.8} & \textbf{78.6} & \textbf{86.6} & 58.2  & \textbf{59.3} & \textbf{66.6} \\
    \bottomrule
    \end{tabular}%
    }}
  \label{tab:comparison_on_shapenet}%
\end{table}%

\subsubsection{\textbf{Performance Comparison on MCB-A Dataset}}
\begin{figure*}[htbp]
  \centering
  \includegraphics[scale=0.5]{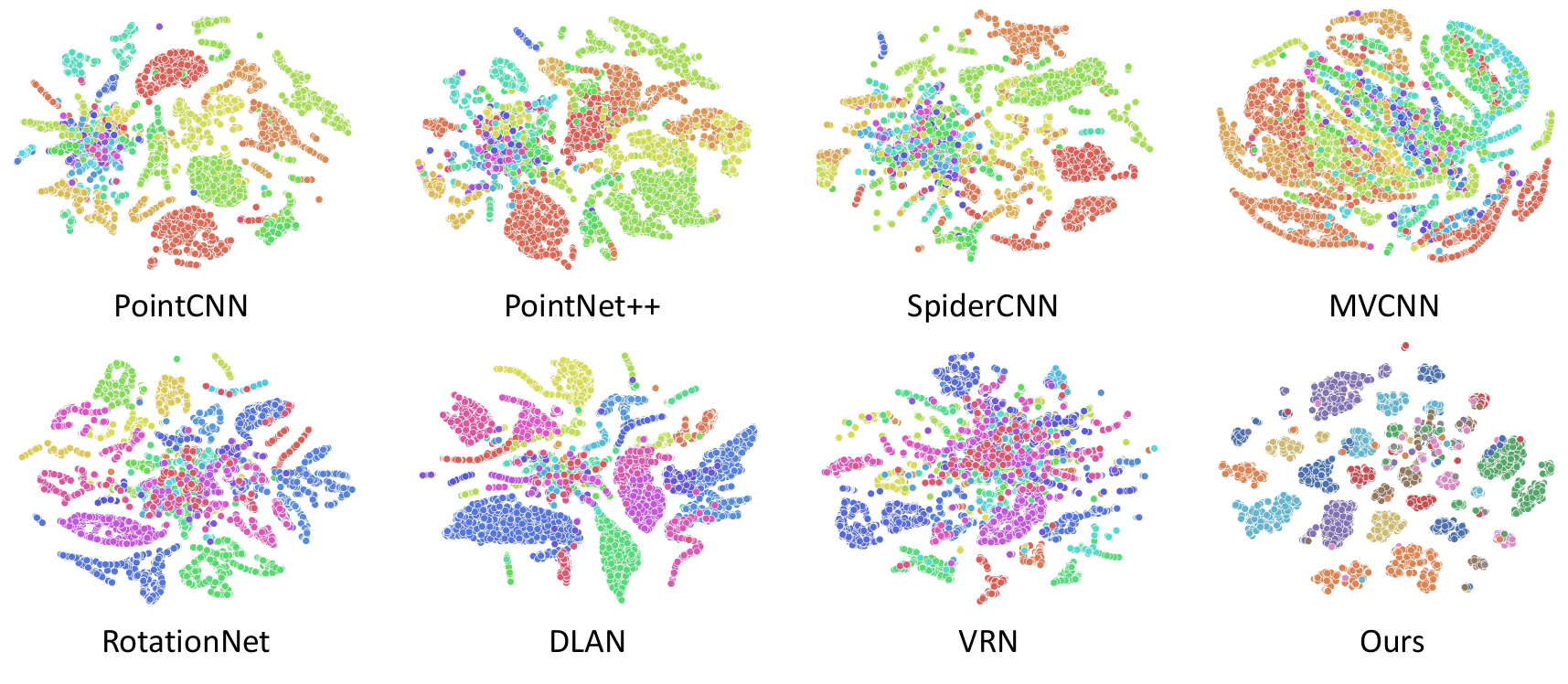}
  \caption{The 2D t-SNE embeddings of testing data in MCB-A dataset.}
  \label{fig:tsne}
\end{figure*}

Table~\ref{tab:comparsions_on_mcb} shows the retrieval performance on MCB-A dataset. We compare our method with those reported in ~\cite{kim2020large}. Among 9 performance metrics, our method secures the first rank in 8 metrics. In particular, on one hand, compared with PointCNN which achieves the best performance among those using only point cloud modality, our method produces 10.4\%, 7.8\% and 10.3\% in F1@N, mAP and NDCG@N under Micro+Macro settings, respectively. On the other hand, compared with RotationNet which achieves the best performance among those using only multi-view modality data, under the Micro+Macro settings, our method achieves scores of 30.6\% in F1@N, 15.1\% in mAP, and 27.0\% in NDCG@N. Additionally, we present the 2D t-SNE embeddings of the testing samples in the MCB-A dataset for all the compared methods in Fig.~\ref{fig:tsne}. The figure illustrates that our method generates the most distinguishable embeddings for the testing samples, leading to the highest retrieval performance.


\begin{table*}[!t]
\caption{Performance comparison on MCB-A dataset (in \%). The best results are in bold.}
\footnotesize
\centering
\scalebox{1.0}{
\begin{tabular}{|c|*{3}{p{1.3cm}<{\centering}}|*{3}{p{1.3cm}<{\centering}}|*{3}{p{1.3cm}<{\centering}}|}
\hline
\multirow{2}{*}{Method} & \multicolumn{3}{c|}{Micro} & \multicolumn{3}{c|}{Macro} & \multicolumn{3}{c|}{Micro + Macro}\\
                         &F1@N &mAP &NDCG@N & F1@N &MAP &NDCG@N & F1@N &mAP &NDCG@N\\ \hline
            PointCNN~\cite{li2018pointcnn} & 69.0 & 88.9 &89.8 & \textbf{83.3} & 88.6 &85.4 &76.2 & 88.3 &87.6\\
            PointNet++~\cite{QiPointNet++} & 61.3 & 79.4 & 75.4 & 71.2 & 80.3 & 74.6 & 66.3 & 79.9 & 75.0\\
            SpiderCNN~\cite{xu2018spidercnn}  & 66.9 & 86.7 & 79.3 & 77.6 & 87.7 & 81.2 & 72.3 & 87.2 & 80.3 \\
            MVCNN~\cite{su2015multi}      & 48.8 & 65.7 & 48.7 & 58.5 & 73.5 & 64.1 & 53.7 & 69.6 & 56.4\\
            RotationNet~\cite{kanezaki2018rotationnet} &50.8 &80.5 &68.3 &68.3 &81.5 &73.5 &56.0 &81.0 &70.9\\
            DLAN~\cite{furuya2015diffusion} &56.8 & 87.9 &82.8 &82.0 &88.0 &84.5 & 69.4 & 88.0 & 83.7\\
            VRN~\cite{brock2016generative} &40.2 & 65.3 & 51.9 & 50.7 & 66.4 & 57.6 & 45.5 & 65.9 & 54.8 \\
            \hline
            Our SCA-PVNet &\textbf{92.8} &\textbf{94.0} &\textbf{99.1} &80.4 &\textbf{98.1} &\textbf{96.7} &
            \textbf{86.6} &\textbf{96.1} &\textbf{97.9}\\
\hline

\end{tabular}
}
\label{tab:comparsions_on_mcb}
\end{table*}

\subsection{Discussion}\label{Discussion}

In this section, we conduct several ablation experiments to analyze the proposed SCA-PVNet.
\subsubsection{\textbf{On the proposed modules}}
We first analyze the effect of the proposed modules. We consider six ablation models: (i) DGCNN using only the point cloud modality; (ii) MVCNN using only multi-view modality; (iii) the model exploiting a direct concatenation of feature representation of (i) and (ii); (iv) the proposed model without view-level feature aggregation branch; (v) the proposed model without object-level feature aggregation branch; (vi) the proposed SCA-PVNet. Table~\ref{tab:ablation_on_proposed_modules} shows the retrieval mAP on ModelNet40 produced by each ablation model. It is clearly observed that the models incorporating both modalities significantly outperforms those with a single modality by around 10\% in mAP. Secondly, it is also shown that the model incorporating the proposed view-level feature aggregation branch or object-level feature aggregation branch outperforms the model which directly concatenates the features generated by DGCNN and MVCNN. Lastly, our SCA-PVNet incorporating both aggregation branches achieves the best retrieval performance among all the ablation methods, showing the effectiveness of the proposed modules.

\begin{table}[htbp]
  \centering
  \caption{Ablation Study on the Proposed Modules}
  \scalebox{0.9}{
    \begin{tabular}{|cc|c|}
    \toprule
    \multicolumn{2}{|c|}{Model}    & mAP \\
    \midrule
    \midrule
    \multicolumn{2}{|c|}{DGCNN (Point Cloud Modality)}     & 81.6 \\
    \multicolumn{2}{|c|}{MVCNN (Multi-View Modality)}    & 80.2 \\
    \multicolumn{2}{|c|}{Direct Concatenation} &91.0 \\
    \midrule
    \multicolumn{2}{|c|}{Ours w/o View-Level Feature Aggregation}  & 91.2 \\
    \multicolumn{2}{|c|}{Ours w/o Object-Level Feature Aggregation}  & 91.4 \\
    \multicolumn{2}{|c|}{Our SCA-PVNet}  & 92.5 \\
    \bottomrule
    \end{tabular}%
  }
  \label{tab:ablation_on_proposed_modules}%
\end{table}%

\begin{figure*}[htbp]
  \centering
  \includegraphics[scale=0.45]{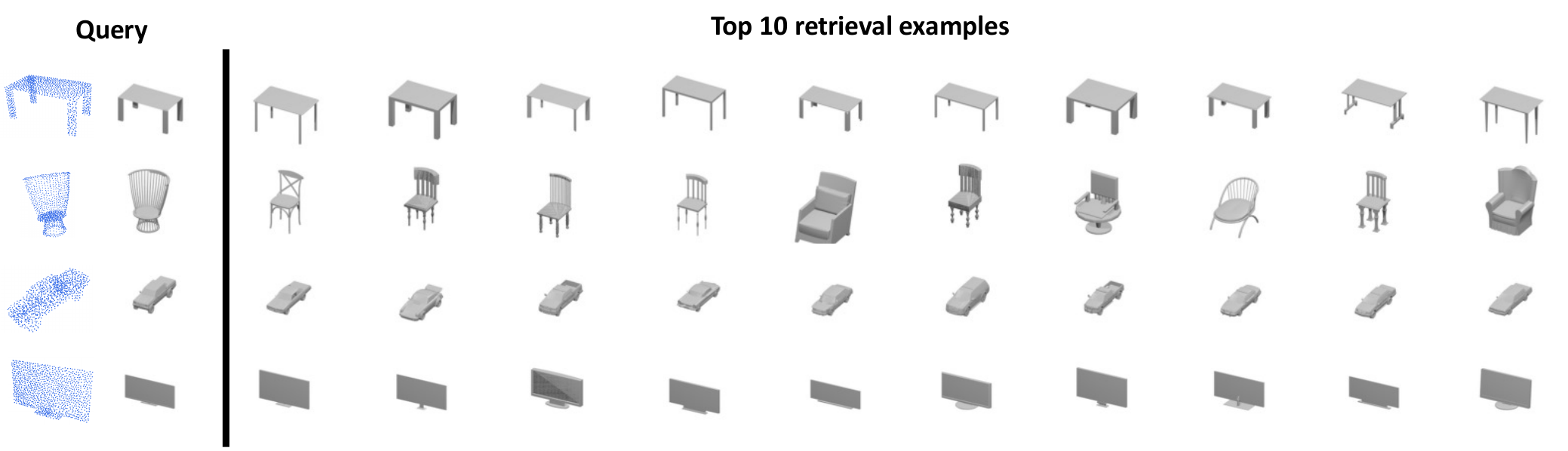}
  \caption{The qualitative results using the testing samples from ShapeNetCore55 dataset.}
  \label{fig:qualitative}
\end{figure*}

\subsubsection{\textbf{On Model Complexity.}}
We also analyze the the relationship between the performance gain and the model complexity. We compare different ablation models of our SCA-PVNet with the baseline method PVRNet~\cite{you2019pvrnet} since PVRNet shares the same feature extraction networks (DGCNN and MVCNN) with our method. Table~\ref{tab:ablation_on_model_complexity} presents the retrieval performance on ModelNet40 and model complexity (in terms of \# of parameters) of four ablation models. It is noted that by progressively incorporating the proposed feature aggregation branches, the performance of our method is consistently improved with only a slight increase in model complexity.

\begin{table}[htbp]
  \centering
  \caption{Ablation Study on Model Complexity}
  \scalebox{0.85}{
    \begin{tabular}{|cc|cc|}
    \toprule
    \multicolumn{2}{|c|}{Model} & mAP   & \# of Params (M) \\
    \midrule
    \midrule
    \multicolumn{2}{|c|}{PVRNet} & 90.5  & 70.000 \\
    \multicolumn{2}{|c|}{Ours w/o View-Level Feature Aggregation} & 91.2  & 70.526 \\
    \multicolumn{2}{|c|}{Ours w/o Object-Level Feature Aggregation} & 91.4 & 74.185 \\
    \multicolumn{2}{|c|}{Our SCA-PVNet} & 92.5 & 75.762 \\
    \bottomrule
    \end{tabular}%
    }
  \label{tab:ablation_on_model_complexity}%
\end{table}%

\subsubsection{\textbf{On Robustness Towards Missing Testing Data}}

In this section, the robustness of the proposed method towards missing testing data is analyzed. In real-world applications, it is common that testing data is only partially available. In our case, it could be missing views or missing points. One merit for multi-modal based approaches is the robustness towards missing data in one of the modality since the other modality could compensate for the information loss for the object to be retrieved.

\begin{figure}[htbp]
  \centering
  \includegraphics[scale=0.38]{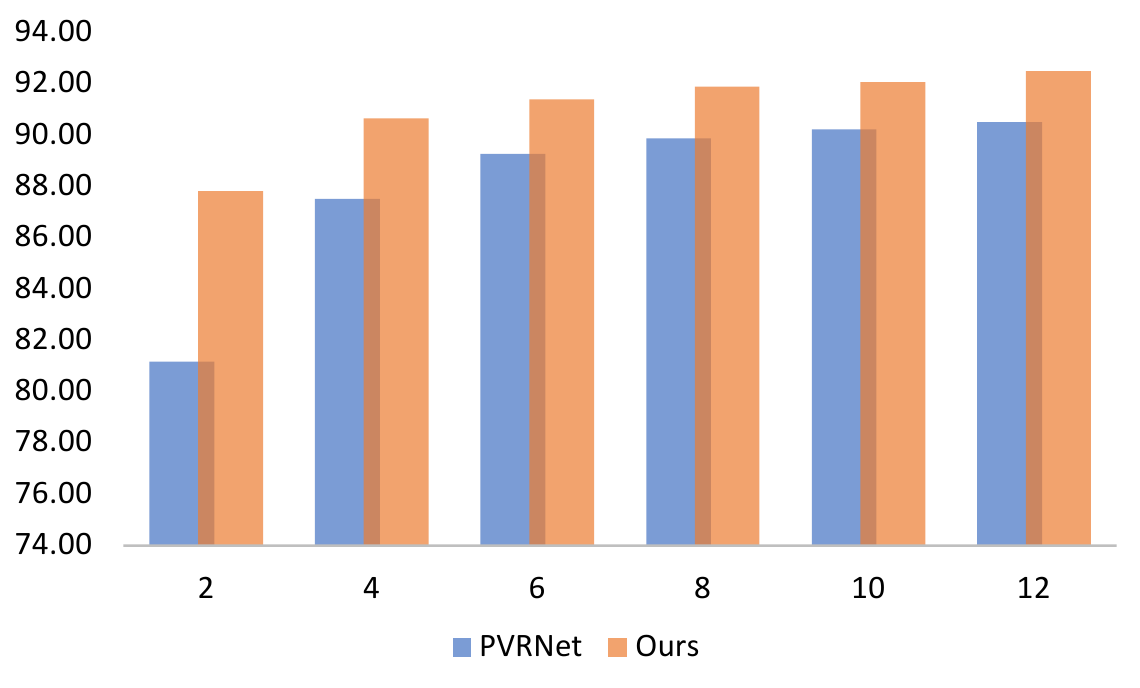}
  \caption{Analysis on missing views.}
  \label{fig:MissingViews_Analysis}
\end{figure}

Motivated by PVRNet~\cite{you2019pvrnet}, we investigate the influence of the scenarios of missing views and missing points, respectively. During training, our SCA-PVNet is trained using 12 views and 1024 points. For missing view scenario, during testing, we retain the number of point for each point cloud as 1024 and vary the number of views to be in the range of [2, 12] with the interval of 2. Fig~\ref{fig:MissingViews_Analysis} shows the mAP with respect to different number of missing views for PVRNet and our method. It is demonstrated that reducing the number of testing views leads to degradation of retrieval performance. By reducing the number of testing views from 12 to 2, compared with the severe performance drop 9.3\% of PVRNet, the performance drop of our method is only 4.7\%. For the edge case of only 2 views are available during testing, our method can still obtain 87.8\% in mAP compared with 81.2\% for PVRNet. In summary, we conclude that our method show better robustness towards missing views compared with PVRNet.

\begin{figure}[htbp]
  \centering
  \includegraphics[scale=0.38]{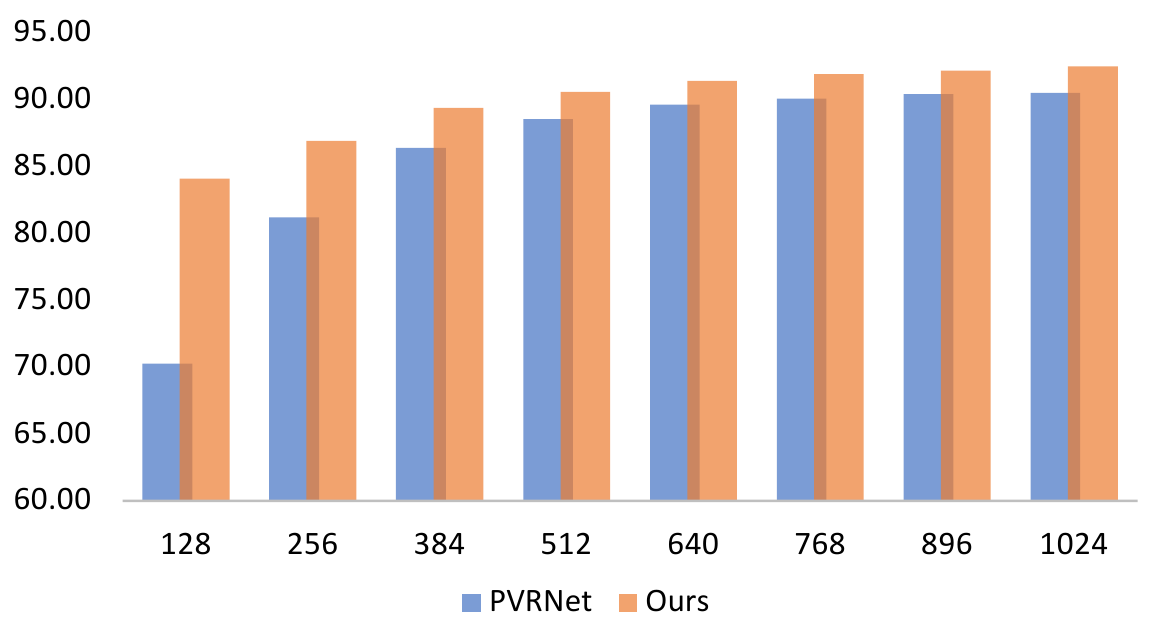}
  \caption{Analysis on missing points.}
  \label{fig:MissingPoints_Analysis}
\end{figure}

In scenarios where points are missing, we keep the number of views at 12 during testing, but the number of points is adjusted within a range of 128 to 1024, increasing in steps of 128. The retrieval mAPs under different settings are illustrated in Fig.~\ref{fig:MissingPoints_Analysis}. Similarly, it is observed that missing points also cause performance degradation. Compared with PVRNet whose retrieval mAP drops from 90.5\% to 70.3\% when reducing the number of points from 1024 to 128, the mAP of our method drops from 92.5\% to 84.1\%. Therefore, we can also conclude that our method is more robust towards missing points.


\subsubsection{\textbf{Qualitative Results}}
In Fig.\ref{fig:qualitative}, we show the top 10 retrieval results for four testing samples (with category ``table'', ``chair'', ``car'' and ``monitor'') from the testing set of ShapeNetCore55. In this illustration, the query sample contains both point cloud and view image modality and we use one view image to represent each retrieved result. From the figure, it is observe that our method can achieve good performance even for those with large within-class variations like the chairs with different styles, as shown in the second row of Fig.\ref{fig:qualitative}.

\section{Conclusion}
In this paper, we have proposed self-and-cross attention based aggregation of point cloud and multi-view images (SCA-PVNet) for 3D object retrieval. To facilitate effective feature fusion, two types of aggregation modules, namely IMAM and CMAM have been proposed by leveraging self-attention and cross-attention mechanism, respectively. Extensive experiments and analysis have been conducted on three datasets, raging from small to large scale, to validate the superiority of the proposed SCA-PVNet over the state-of-the-art methods. In the future, we will further investigate how to extend the current self-and-cross attention framework to more representation modalities of 3D objects, such as voxels or panorama-view images.


\bibliographystyle{IEEEtran}
\bibliography{tmm_main}
%
\end{document}